# KEYPOINT-BASED OBJECT TRACKING AND LOCALIZATION USING NETWORKS OF LOW-POWER EMBEDDED SMART CAMERAS

Ibrahim Abdelkader[1], Yasser El-Sonbaty[1] and Mohamed El-Habrouk[2]
[1]Dept. of Computer Science, Arab Academy for Science & Technology, Alexandria, Egypt
[2]Dept. of Electrical Engineering, Faculty of Engineering, Alexandria, Egypt

**ABSTRACT**

Object tracking and localization is a complex task that typically requires processing power beyond the capabilities of low-power embedded cameras. This paper presents a new approach to real-time object tracking and localization using multi-view binary keypoints descriptor. The proposed approach offers a compromise between processing power, accuracy and networking bandwidth and has been tested using multiple distributed low-power smart cameras. Additionally, multiple optimization techniques are presented to improve the performance of the keypoints descriptor for low-power embedded systems.

**KEYWORDS**

WSNs, Object Tracking, ORB, Machine Vision, Distributed Cameras

## 1. INTRODUCTION

Object localization and tracking has been the subject of many research studies, with applications in military, surveillance, robotics and more. This paper introduces a new keypoints-based approach for object tracking and localization, which offers a compromise between processing power, detection rate and network bandwidth. Additionally, the paper presents optimization techniques to allow the keypoints descriptor to run on low-power smart cameras. The proposed approach uses a multi-view keypoints descriptor to track objects, created by fusing together multiple descriptors of the same object extracted from different views as shown in Figure 1. Each camera performs simple background subtraction, locally, via frame differencing to detect moving objects. The size and location of detected objects are used as the ROIs (Region of Interests) for keypoints extraction. Instead of sending full frames, each camera only sends the extracted keypoints to a server which performs the matching against the multi-view descriptor. Since the locations of the cameras are assumed to be fixed, object locations can be estimated as well.

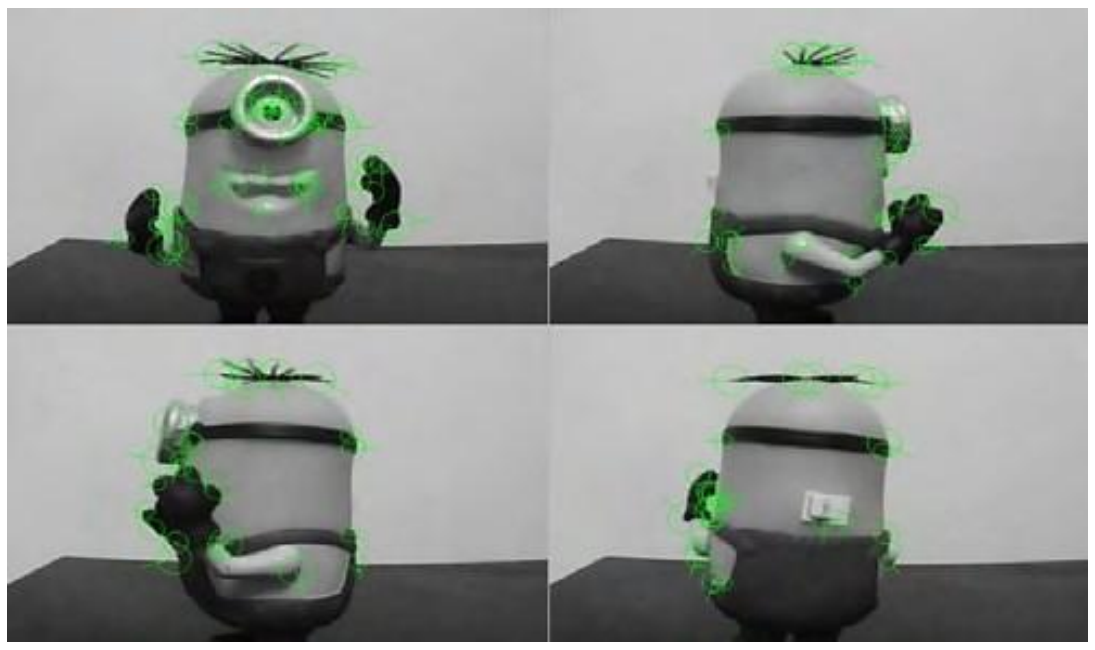

Figure 1. Multi-view keypoints descriptor

## 2. RELATED WORK

Many different approaches with different end applications have been proposed for object tracking and localization, however a full survey of object tracking is beyond the scope of this paper, thus only the relevant work is reviewed in this section.

(Ercan et al, 2007) proposed a camera network to track a single object in the presence of static and moving occluders. Camera locations are assumed to be fixed. Additionally full knowledge about static occluders is required and some information about moving occluders are also assumed to be known. Simple background subtraction is performed on each node and the object is detected using features. The scene is then reduced to a single scan line which is sent to a cluster head for further processing.

(Yang et al, 2003) used a camera network to count people for the purposes of surveillance. Each camera performs local processing to extract foreground objects from the background, and sends bitmaps over the network for further processing. In contrast to our work, full bitmaps are sent over the network which increases network bandwidth and consumption.

(Shen et al, 2012) proposed an efficient background subtraction technique to implement an object tracking system using an embedded camera network. The background subtraction algorithm is shown to be as accurate as traditional systems yet faster, however the system can only track a single object and it is based on background subtraction which is affected by different lighting conditions.

(Nummiaro et al, 2003) described a multi-view color-based object tracking system for tracking faces using color histograms and particle filters. The system is robust against partial occlusions and transformations, however it depends on fixed lighting conditions.

(El-Sonbaty & Ismail, 2003) proposed a graph-based algorithm for matching partially occluded objects. The algorithm matches connected lines based on distance ratio. The algorithm is translation, rotation and scale invariant, however its complexity does not allow it to run on low-power, real-time systems, although the algorithm can be accelerated via parallelism (Nagy et al, 2006) or compression (El-Sonbaty et al, 2003).

(Rublee et al, 2011) created ORB (Oriented FAST and Rotated BRIEF), a fast binary feature descriptor. ORB is based on the BRIEF (Calonder et al, 2010) descriptor, designed to be a more efficient alternative to SIFT (Lowe, 2004) and SURF (Bay et al, 2006). The ORB descriptor is computed using simple binary tests between the pixels in the patch around a keypoint. To produce multi-scale keypoints, the FAST (Rosten & Drummond, 2006) corner detector is used on an image pyramid. The keypoints are sorted by Harris (Harris & Stephens, 1988) corner score and the top keypoints are selected. To achieve rotation-invariance, the orientation of each keypoint is determined using the *intensity centroid* (Rosin, 1999) and used to rotate (or steer) the binary sampling pattern.

The proposed approach uses FAST and ORB for feature extraction and description respectively. Using ORB makes the system inherently scale, rotation and lighting invariant and robust against partial occlusions. However, while ORB is much faster than alternatives (SURF or SIFT) improvements to the performance of the tracker were needed to allow it to run on a low-power embedded system. The following section presents ORB a few optimization techniques implemented in the tracking system.

## 3. OPTIMIZING ORB

To extract multi-scale descriptors, the FAST corner detector runs on all image pyramid levels. However, when matching this descriptor, the corner detector is only run on the first scale (full image). Since the descriptor contains keypoints from different scales, a detected keypoint will have a match in one of the scales. A similar approach is used in the work done by (Wagner et al, 2008) to optimize the SIFT descriptor for mobile phones. Additionally, the descriptor size is limited to 32-bytes, which performs almost as good as a 64-bytes descriptor as shown in Figure 2 from the experiments performed by (Calonder el al, 2010).

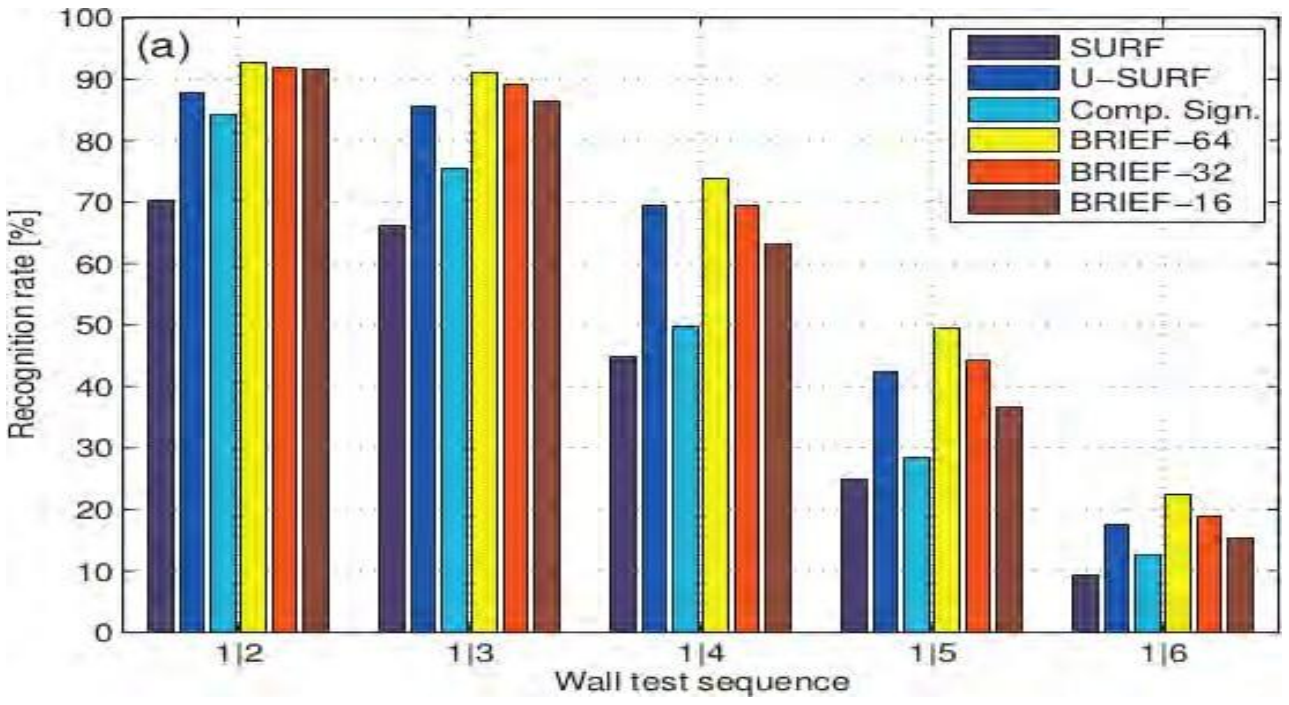

Figure 2. BRIF recognition rate using 16, 32 and 64 bytes

Furthermore, the keypoints descriptor is optimized by removing redundant keypoints detected at multiple octaves at the same position with the same or close angles of rotation. For example, keypoints which appear at multiple octaves at the same location and have the same (or slightly different) rotation angles, are merged into one keypoint in the final descriptor. Since ORB requires a Gaussian smoothing at every scale, a separable convolution (horizontal convolution followed by a vertical one) is used instead of a 2D convolution, to lower the number of operations required to perform the smoothing. Additionally, SIMD instructions are used in the inner loop of the separable onvolution to speed up the pixel level operations.

# 4. EXPERIMENTS AND RESULTS

The platform used for evaluation is a low-power machine vision camera developed by the author, called OpenMV shown in Figure 3-a. OpenMV is based on the STM32F7 ARM Cortex-M7 dual-issue MCU running at 216MHz. The MCU has 512KiBs SRAM, 2MiBs flash, a single precision FPU and DSP instructions. In addition, it includes a digital camera interface (DCMI), JPEG encoder and multiple serial peripheral interfaces. The camera can be extended with BLE (Bluetooth Low Energy) or WiFi modules, and consumes about 500mW on average when doing image processing.

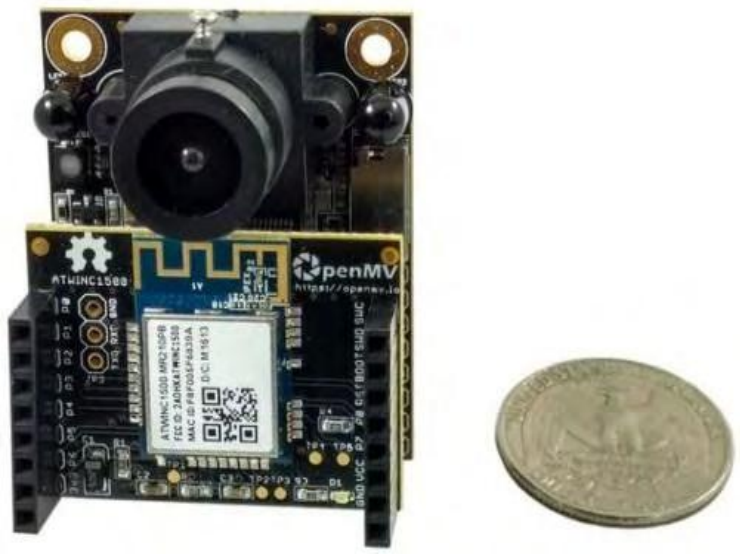

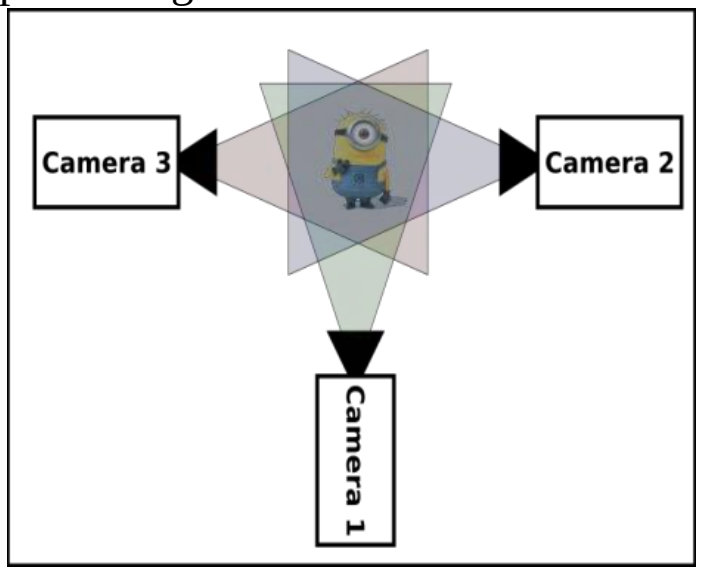

(a) (b)

Figure 3. (a) OpenMV camera with WiFi module. (b) An illustration of the Camera network setup

Multiple OpenMV cameras are placed in fixed locations to monitor a scene. Each camera is equipped with a battery for power supply and a 2.4GHz WiFi/802.11 extension module for communications. The cameras can use BLE for communications instead to reduce the power consumption further. Additionally BLE RSSI can be used to determine the distance between cameras without assuming prior knowledge of their locations. When a moving object is detected, the camera uses the object's location and size as the ROI passed to Fast/ORB to extract keypoints. The ROI and extracted keypoints descriptors are sent over the network to a sever for further processing. See Figure 3-b for an illustration of the setup used in this experiment.

A total of 38 bytes is sent for each keypoint (4 bytes for its location, 2 bytes for the angle of rotation and a 32 bytes descriptor). Limiting the maximum number of keypoints to 50 keypoints per frame, each camera sends a maximum of 1900 bytes (~1.8KiBs) for detection compared to sending a full frame (300KiBs at 640x480 grayscale). Figure 4 shows an example from the first camera, starting with frame differencing to detect moving objects followed by background subtraction, keypoints detection and keypoints optimization to further reduce the number of detected keypoints. The two other cameras produce similar results with different ROIs sizes depending on the distance of the object from each camera.

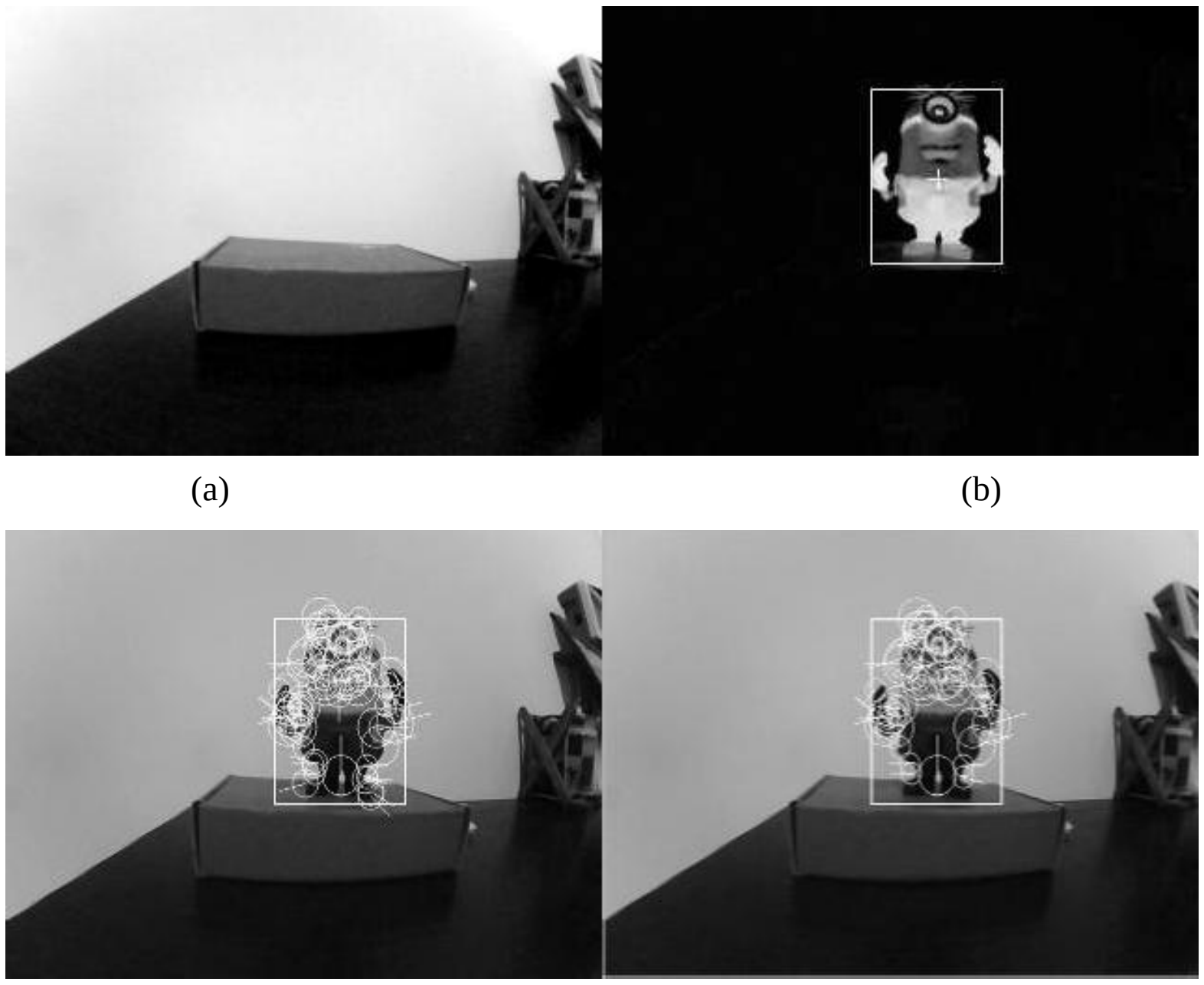

(a) (b)

(c) (d)

Figure 4. An overview of the approach. (a) The background image (b) moving object detection with frame difference (c) ROI used to detect keypoints (d) Keypoints reduction after optimization

The same scene is observed by two more cameras, which perform the same tasks and send their detection results to a server for further processing. The server performs object matching using the multi-view keypoints descriptor and localization using the ROI sizes. Figure 5 shows tracking results plotted compared to the ground truth.

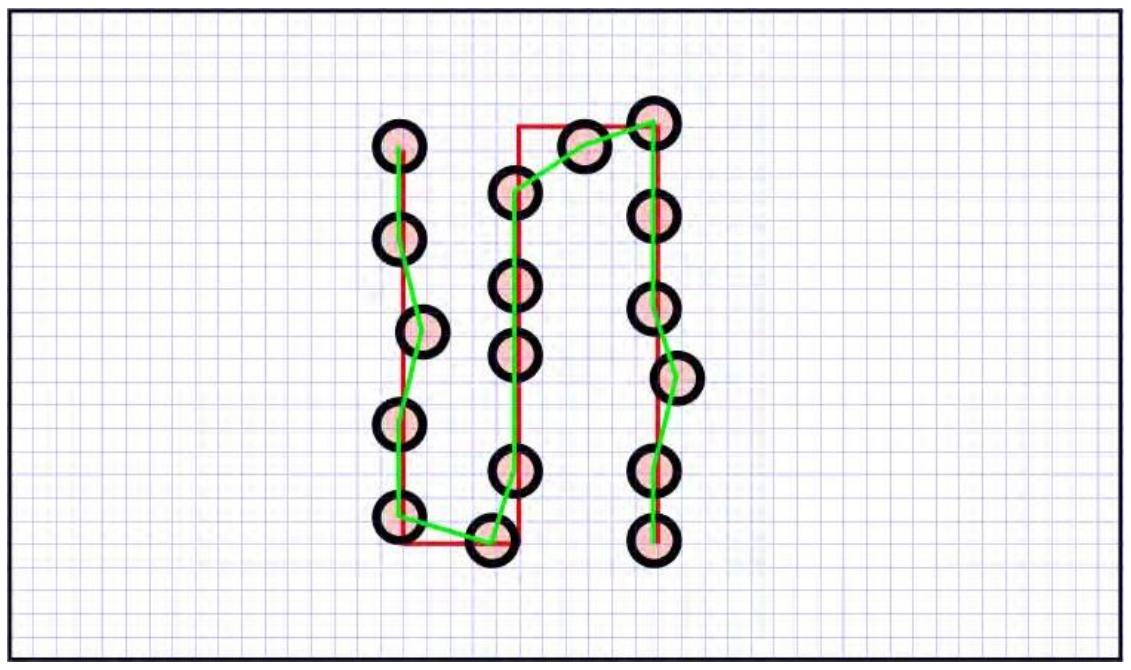

Figure 5. The tracking results. The circles represent detection locations connected with a green line. The ground truth is plotted in red

## 5. CONCLUSION

This paper presented a new approach for object tracking and localization. The proposed approach offers a compromise between processing power and networking bandwidth by sending keypoints instead of full frames, reducing the communications overhead significantly. Each camera sends only the location of the detected object and extracted keypoints to a server which performs the object matching using a multi-view keypoints descriptor. The size of the detected object as well as angles of rotations of matched keypoints allow the object location and orientation to be estimated as well. This paper also presented multiple optimization techniques to allow ORB to run on a low-power smart camera.